\documentclass[letterpaper, 10 pt, conference]{ieeeconf}

\IEEEoverridecommandlockouts

\usepackage{color}
\usepackage{multirow}
\usepackage{booktabs}
\usepackage{url}
\usepackage{subfigure}
\usepackage{threeparttable}
\usepackage{cite}
\usepackage{hyperref}[colorlinks,linkcolor=blue]
\usepackage[ruled,linesnumbered]{algorithm2e}
 \renewcommand{\baselinestretch}{0.966}

\makeatletter
\def\@citex[#1]#2{\leavevmode
\let\@citea\@empty
\@cite{\@for\@citeb:=#2\do
{\@citea\def\@citea{,\penalty\@m\ }%
\edef\@citeb{\expandafter\@firstofone\@citeb\@empty}%
\if@filesw\immediate\write\@auxout{\string\citation{\@citeb}}\fi
\@ifundefined{b@\@citeb}{\hbox{\reset@font\bfseries ?}%
\G@refundefinedtrue
\@latex@warning
{Citation `\@citeb' on page \thepage \space undefined}}%
{\@cite@ofmt{\csname b@\@citeb\endcsname}}}}{#1}}
\makeatother

\usepackage{enumerate}
\usepackage{amssymb}
\usepackage{leftidx}
\usepackage{amsfonts}
\usepackage{amsmath}
\usepackage{bm}
\usepackage{booktabs}
\usepackage{setspace}
\usepackage{makecell}
\usepackage{setspace}
\usepackage{float}
\usepackage{subfigure}
\usepackage[pdftex]{graphicx}

\usepackage{tikz}
\usetikzlibrary{shapes,snakes}

\graphicspath{{figs/}}
\usepackage{graphicx}
\graphicspath{{figs/}}
\newcommand{\etal}{\textit{et al}.}
\newcommand{\ie}{\textit{i}.\textit{e}.}
\newcommand{\eg}{\textit{e}.\textit{g}.}
\begin{document}
\thispagestyle{empty} 
\pagestyle{empty}  

\title{\LARGE \bf
Bubble Planner: Planning High-speed Smooth Quadrotor Trajectories using Receding Corridors
}
\author{Yunfan~Ren*, Fangcheng~Zhu*, Wenyi Liu, Zhepei Wang, Yi Lin, Fei Gao and
 Fu~Zhang
 \thanks{*These two authors contributed equally to this work.}
 \thanks{Y. Ren, F. Zhu, and F. Zhang are with the Department of Mechanical Engineering, University of Hong Kong \texttt{\{renyf, zhufc\}@connect.hku.hk, fuzhang@hku.hk}, W. Liu is with School of Electronics and Information Engineering, Harbin Institute of Technology, Shenzhen \texttt{180210215@stu.hit.edu.cn}, Z. Wang and F. Gao are with the College of Control Science and Engineering, Zhejiang University \texttt{\{wangzhepei, fgaoaa\}@zju.edu.cn} Y. Lin is with Dji Co. \texttt{ylinax@connect.ust.hk}.}
}
\maketitle

\begin{tikzpicture}[overlay, remember picture]
  \path (current page.north) ++(0.0,-1.0) node[draw = black] {Accepted for the 2022 IEEE/RSJ International Conference on Intelligent Robots and Systems (IROS), Kyoto, Japan};
\end{tikzpicture}
\vspace{-0.3cm}

\pagestyle{empty}  
\thispagestyle{empty} 

\begin{abstract}
 Quadrotors are agile platforms. With human experts, they can perform extremely high-speed flights in cluttered environments. However, fully autonomous flight at high speed remains a significant challenge. In this work, we propose a motion planning algorithm based on the corridor-constrained minimum control effort trajectory optimization (MINCO) framework. Specifically, we use a series of overlapping spheres to represent the free space of the environment and propose two novel designs that enable the algorithm to plan high-speed quadrotor trajectories in real-time. One is a sampling-based corridor generation method that generates spheres with large overlapped areas (hence overall corridor size) between two neighboring spheres. The second is a \textit{Receding Horizon Corridors} (RHC) strategy, where part of the previously generated corridor is reused in each replan. Together, these two designs enlarge the corridor spaces in accordance with the quadrotor's current state and hence allow the quadrotor to maneuver at high speeds. We benchmark our algorithm against other state-of-the-art planning methods to show its superiority in simulation. Comprehensive ablation studies are also conducted to show the necessity of the two designs. The proposed method is finally evaluated on an autonomous LiDAR-navigated quadrotor UAV in woods environments, achieving flight speeds over $13.7 m/s$ without any prior map of the environment or external localization facility. 
\end{abstract}


\section{Introduction}
Quadrotors are proved to be one of the most agile platforms which perform increasingly complex missions in different scenarios. However, high-speed flight in unknown environments is still an open problem. The limits on payload and onboard sensing make this task especially challenging for aerial robots \cite{tordesillas2021faster}. To achieve high-speed flights, trajectory planning is of vital importance to ensure the safety (i.e., collision avoidance \cite{huang2019collision}), smoothness, and fast maneuvers facing unknown obstacles.

High-speed trajectory planning in unknown environments is a great challenge, especially in the replanning phase where the high quadrotor speeds require extremely agile maneuvers to avoid newly-sensed obstacles. Existing (re-)planning methods \cite{zhou2019robust, liu_sfc,tgk_planner} typically consist of a frontend that aims to find a guiding path (or flight corridor) and a backend that smooths the trajectory around the guiding path (or optimizes a smooth trajectory within the corridor). The main difficulty in this framework is how to design the frontend such that the replanned guiding path (or flight corridor) is feasible: at least one dynamically-feasible and obstacle-free solution can be found in the backend optimization. A poorly-designed frontend may leave too little space for the quadrotor to avoid obstacles (\eg, decelerate or make turns), hence leaving no dynamically feasible solution in the subsequent trajectory optimization. Another difficulty is the backend optimization, which needs to perform both temporal and spatial deformation in an efficient manner such that the maximal speed can be attained. 

\begin{figure}[t]
 \centering 
 \includegraphics[width=0.40\textwidth]{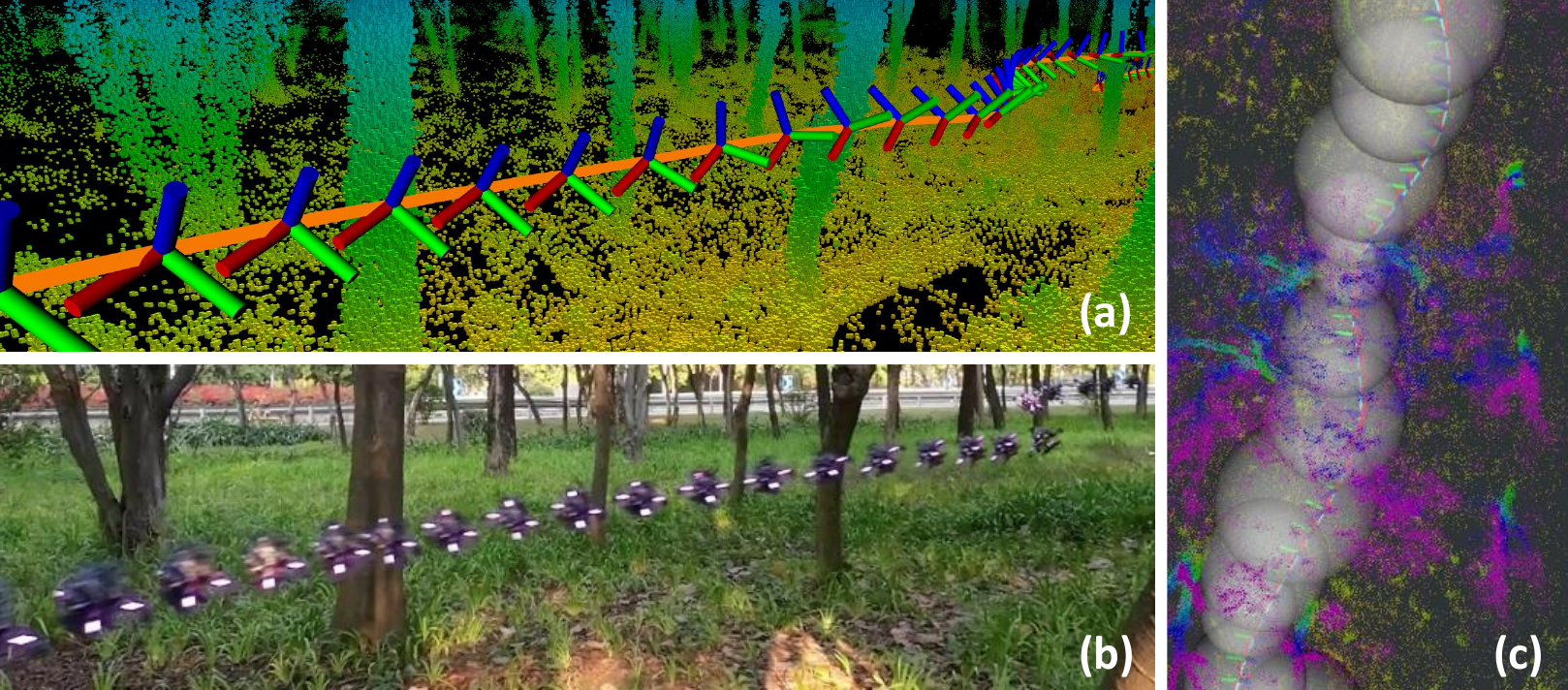}
 \caption{High-speed navigation in the wild. (a) The generated point cloud map during the flight. (b) Composite images of the same flight. (c) The generated sphere-shaped flight corridor. Video is available at \href{https://youtu.be/7tQCV6KBzSY}{https://youtu.be/7tQCV6KBzSY} }
 \label{fig:fig1}
\end{figure}

\begin{figure}[t]
 \centering 
 \includegraphics[width=0.4\textwidth]{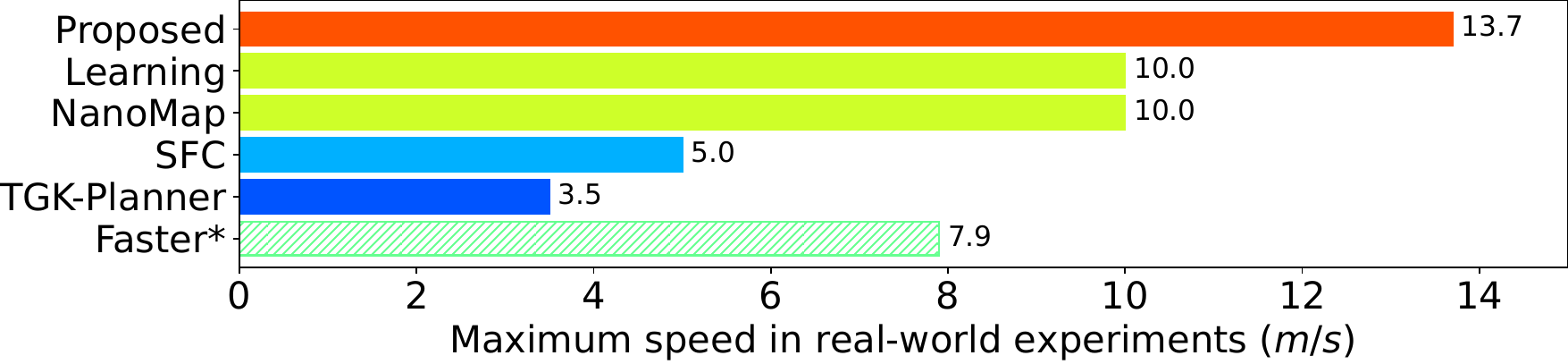}
 \caption{Comparison of the maximum speed in real-world experiments (the flight speed of other methods are read from their original papers). The {Faster*} \cite{tordesillas2021faster} baseline uses a motion capture system as state feedback and a depth camera to detect obstacles. The proposed method and {SFC} \cite{liu_sfc} use a LiDAR with IMU for navigation while {NanoMap} \cite{florence2018nanomap}, {Learning} \cite{loquercio2021learning}, {TGK-Planner} \cite{tgk_planner} use a RGB-D camera with IMU. All of the above methods except Faster are performed in real-world forests, and Faster is tested in an indoor artificial environment. Our approach reaches a maximum speed of over $13.7~m/s$.}
 \label{fig:real_vel}
\end{figure}

In this paper, we propose a robust and efficient motion planning algorithm to address the above issues systematically. The overall algorithm is based on a corridor approach. In the backend, we adopt a state-of-the-art minimum control effort optimization (MINCO) framework \cite{wang2022geometrically} to deform the trajectory temporal and spatial parameters efficiently. Our contribution in this paper mainly lies in the frontend, including:

\begin{itemize}
 \item [1)] 
 A novel sampling-based corridor generation method that preserves large corridor volume by considering the size of each sphere and their overlapped spaces. The increased corridor volume allows more space for the quadrotor to maneuver (hence succeed) at high speeds.
 \item [2)]
 A \textit{Receding Horizon Corridors} (RHC) scheme that reuses corridors in the previous planning cycle. Specifically, in each replan, the first part of the flight corridor is directly from the previous planning cycle, and the second part is generated according to newly-sensed obstacles. This receding scheme ensures the corridor in each replan always contains sufficient space for the quadrotor to maneuver from its current state, significantly improving the replan process's success rate and convergence speed under high-speed flight.
 \item [3)]
 A real-time planning system that integrates these two designs of frontend with the MINCO backend \cite{wang2022geometrically}. A comprehensive benchmark comparison and an ablation study are conducted in simulation to show the superiority of our system and the effectiveness of the two designs.
 
 \item [4)]
{Implementation and validation} the proposed method on a fully autonomous quadrotor system. Multiple real-world tests show that our methods achieve flight speeds over $13.7~m/s$ (see Fig. \ref{fig:fig1}).

\end{itemize}

\section{Related Works}

\subsection{High-Speed Navigation in the Wild}

Various approaches have been proposed to enable autonomous quadrotor flights in unknown environments. Florence \etal~ \cite{florence2020integrated} propose a reactive planner, which takes depth image as input and selects the best trajectory from a pre-built motion primitives library. The work in \cite{florence2018nanomap} proposes an uncertainty-aware lazy search map called NanoMap on the reactive controller and achieves a maximum flight speed of $10~m/s$. Although it has a low computation complexity, the pre-built set of motion primitives is relatively small, making it difficult to cover fine maneuvering skills that are necessary when the quadrotor is facing new, unexpected obstacles during high-speed flights. Similar motion primitive-based method is used (as a frontend) by Zhou \etal~\cite{zhou2019robust}, Liu \etal~\cite{liu2017search}, Zhang \etal~\cite{zhang2020falco} and Kong \etal~\cite{kong2021avoiding}, which therefore suffer from similar drawbacks. Ye \etal~\cite{tgk_planner} utilizes a frontend based on RRT* kinodynamic sampling. Similar to the motion primitive methods, the sampled states are usually in low dimensions (\eg, position and velocity) and few in numbers in order to ensure sufficient computation efficiency, making it very difficult to produce fine quadrotor maneuvers in high-speed flights. Unlike the previous methods \cite{tgk_planner,zhou2019robust,liu_sfc}, which typically have a frontend planning a rough path from the quadrotor's current position to the target one and a backend which further refines the trajectory by optimization, Zhou \etal~\cite{zhou2020ego} proposed to plan a whole trajectory without considering any obstacle in the first stage and then locally modify the trajectory to fly around the detected obstacles. The local trajectory modification is achieved efficiently by directly incorporating a repulsive force from obstacles in the trajectory optimization cost function. The repulsive force is similar to a coarse-level distance field and hence suffers from the local minimum problem, hence unsuitable for high-speed trajectory planning. Another interesting method is proposed by Loquercio \etal~\cite{loquercio2021learning}, they use imitation learning to generate a trajectory directly from the depth image and current state. Limited by the sensing range and noise, the success rate of their methods decreases when forward speed is over 10 $m/s$. Compared with the methods mentioned above, our method achieves much higher flight speed in both simulation and experiments (see Fig. \ref{fig:real_vel}).

\subsection{Corridor-based Trajectory Planning}

Corridor-based trajectory planning methods, which use geometrical shapes to represent free space, have been popular in recent years. Chen \etal~\cite{chen2015real} build a discrete graph from an OctoMap structure \cite{hornung2013octomap} and directly {use} free cubes in OctoMap as the corridor constraints. Liu \etal~\cite{liu_sfc} use polyhedrons to represent the free space, also called convex decomposition. Each cube or polyhedron on the flight corridor then imposes multiple linear hyperplane constraints in the subsequent trajectory optimization. {Sphere-shaped} corridors are also very commonly used. Compared with polyhedrons, a sphere imposes only one constraint in the trajectory optimization. It can often be quickly obtained by \textit{Nearest Neighbor Search} (NN-Search) using a KD-Tree structure. Gao \etal~\cite{gao2019flying} propose a sphere-shaped corridor generation scheme under the RRT* framework. Ji \etal~\cite{ji2021mapless} propose a forward-spanning-tree-based spherical corridor generation scheme. These two methods can generate corridors in a relatively short time. However, their corridor generation process only considers the connectivity of adjacent spheres. The found spheres often have small overlaps between adjacent ones, which over constrains the subsequent trajectory optimization and leaves tiny space for the quadrotor to maneuver at high speeds. Another problem is the lack of explicit consideration of the quadrotor's current speed, the resultant flight corridor often does not contain sufficient space for the quadrotor to maneuver from its current speed. The two problems will considerably reduce the feasible solution space and cause the backend optimization to fail. In contrast, our frontend attempts to find large individual spheres and their overlaps, while the receding scheme automatically incorporates the quadrotor current speed in each replan. These two designs greatly improve the success rate and convergence speed of the subsequent trajectory optimization.

Trajectory optimization with the corridor constraint is also well studied by some recent works. Ji \etal~\cite{ji2021mapless} use an alternating minimization method \cite{am_traj} and iteratively {insert} waypoints to ensure that the trajectory completely falls in the corridor. However, the waypoints are selected heuristically, which leads to sub-optimal solutions. Mellinger \etal~\cite{mellinger2011minimum} use piece-wise polynomial to represent the trajectory and generate a minimum-snap trajectory by solving a quadratic programming (QP) problem. The corridor constraints are used as inequality constraints in the QP. Gao \etal~\cite{gao2019flying} use B-spline to represent trajectories and formulate the corridor constraints and trajectory optimization into a second-order cone programming (SOCP) problem. Both methods solve the optimization problem with hard constraints and have quite significant computation time. Our approach is most similar to \cite{wang2022geometrically}. {The corridor constraints are first eliminated by a $C^2$-continuous barrier function. } Then, a spatial-temporal deformation is performed. The optimization problem is finally turned into an unconstrained one that can be solved by Quasi-Newton methods efficiently and robustly.

\section{Preliminaries}

\label{sec:minco}

\begin{figure}[t]
 \centering 
 \includegraphics[width=0.4\textwidth]{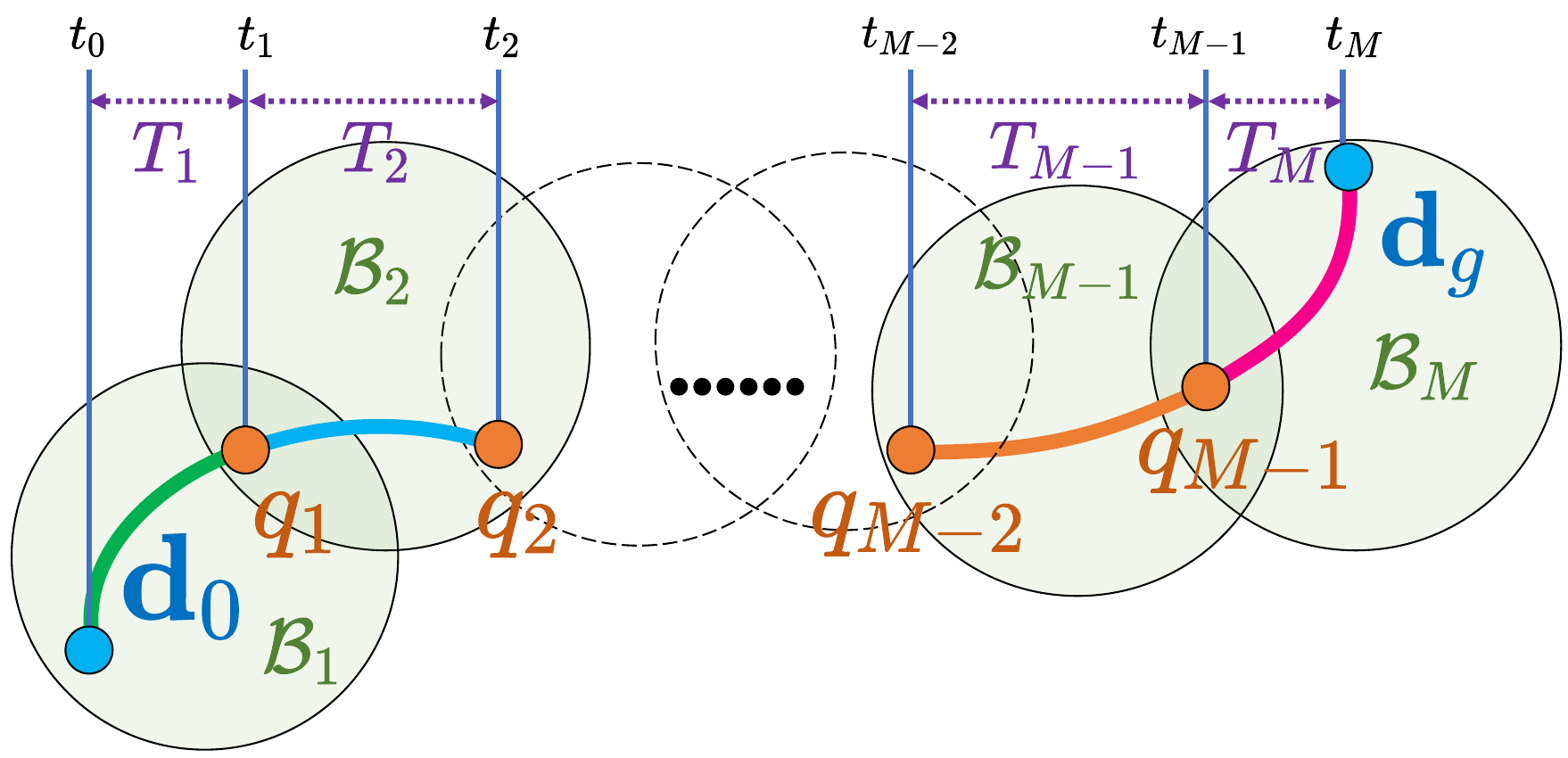}
 \caption{The whole trajectory is composed of $M$ pieces, contained in their respective sphere. The green areas $\mathcal B_i$ are the spheres. The orange points $q_i $ are the intermediate waypoints, which are always constrained in the intersecting space of two adjacent spheres. $T_i$ is the time allocation of each piece. $\mathbf d_0,\mathbf d_g$ are the given initial and goal states.} 
 \label{fig:poly_traj}
\end{figure}

In this section, we briefly go through the backend trajectory optimization used in our algorithm. We model the quadrotor to a non-linear dynamic system following \cite{mellinger2011minimum}, which is proved to be differential flat with flat output $\sigma = [x,y,z,\psi]^T$ with
 $p = [x,y,z]^T$ the quadrotor position in the world frame and $\psi$ the yaw angle. Due to the differential flatness, it is sufficient to plan the flat output trajectory $\sigma(t)$. In this work, we only plan the position trajectory $p(t)$ and specify the yaw angle trajectory $\Phi(t)$ as the tangent direction of $p(t)$ such that the quadrotor is always facing forward during a flight.

As shown in Fig. \ref{fig:poly_traj}, given a flight corridor $\mathcal B$ that consists of a sequence of overlapping spheres (each is denoted by $\mathcal{B}_i, i = 1, \cdots, M$, see Sec.~\ref{sec:sfc}), the goal of the trajectory optimization is to find a smooth trajectory $p(t): [0, t_M] \mapsto \mathbb{R}^3$ over time duration $t_M$ that connects the initial position $q_0 \in \mathbb{R}^3$ at time zero to the terminal one $q_M \in \mathbb{R}^3$ at time $t_M$ and is completely contained in the sphere-shaped corridor $\mathcal B$. 

In practice, the smoothness of the trajectory is quantitatively represented by the magnitude of its $s$-th order derivative $ \| p^{(s)}(t) \|_2^2$ ($s=4$ in experiments). Moreover, the trajectory $p(t)$ is usually decomposed into $M$ pieces, each piece $p_i(t)$ is contained in sphere $\mathcal{B}_i$ for the time period $t \in [t_{i-1}, t_i]$, \ie, 
\begin{equation}
 p(t) = p_i(t - t_{i-1}) \in \mathcal{B}_i, t \in [t_{i-1}, t_i] 
\end{equation}

Adjacent trajectory pieces $p_{i}(t)$ and $p_{i+1}(t)$ should meet at the same point $q_i \in \mathbb{R}^3$ at time $t_i$. Moreover, the trajectory $p(t)$ should start at a given initial state $\mathbf{d}_0$ (up to $(s-1)$-th order derivative) and terminate at a given goal state $\mathbf{d}_g$ (up to $(s-1)$-th order derivative). Considering these constraints and kinodynamic constraints (\eg, speed and acceleration), the trajectory optimization can be formulated as 
\begin{subequations}\label{eq:optimization}
 \begin{alignat}{4}
 \min\limits_{p(t)} & \int_{0}^{t_{M}} \| p^{(s)} (t) \|_2^2 dt + \rho_T t_M \label{eqa:obj_a}\\ 
 s.t.~~&p^{(0:s-1)}(0) = \mathbf{d}_0, p^{(0:s-1)}(t_M) = \mathbf{d}_g \label{eqa:obj_b}\\
 & p(t_i) = q_i, \forall 1 \leq i < M \label{eqa:obj_c} \\ 
 & t_{i-1} < t_i, \forall 1 \leq i \leq M \label{eqa:obj_d} \\
 &\|p^{(1)}(t) \|_2^2 \leq v_{max}^2, \|p^{(2)}(t) \|_2^2 \leq a^2_{max}, \label{eqa:obj_e} \\
 & p(t) = p_i(t-t_i) \in \mathcal{B}_i, \forall 1\leq i \leq M, t \in [t_{i-1}, t_i] \label{eqa:obj_f}
 \end{alignat}
\end{subequations}
where $\rho_T$ is the weight penalizing the total trajectory time $t_M$ such that the maximal allowed speed $v_{max}$ can be attained. 

The optimization in (\ref{eq:optimization}) can be solved in two steps: in the first step, we fix the intermediate way point $\mathbf q = (q_1,\dots,q_{M-1})$ and time allocation vector $\mathbf T = (T_1,\dots,T_{M})$, where $T_i \triangleq t_{i} - t_{i-1} > 0$, and optimize only the first part (\ie, the smoothness) of (\ref{eqa:obj_a}) considering only the constraints in (\ref{eqa:obj_b}) and (\ref{eqa:obj_c}). Shown in \cite{wang2022geometrically}, this optimization problem leads to an optimal solution where each piece $p_i(t)$ is a $(2s-1)$-th order polynomial and its coefficients are uniquely determined from $(\mathbf q, \mathbf T)$, \ie,
\begin{equation}\label{eq:minco_trajectory}
 p_i(t) = \mathbf c_i(\mathbf q, \mathbf T)^T \beta(t), t \in [0, T_i] 
\end{equation}
where $\mathbf c_i \in \mathbb R^{2s\times 3}$ is the coefficient matrix depending on $(\mathbf q, \mathbf T)$ and $\beta(t)= [1 ,t,\dots , t^{2s-1}]^T$ is the time basis function. 

In the second step, the complete problem in (\ref{eq:optimization}) is optimized from the class of trajectories parameterized in (\ref{eq:minco_trajectory}). Since the trajectory in (\ref{eq:minco_trajectory}) naturally satisfies the constraints in (\ref{eqa:obj_b}) and (\ref{eqa:obj_c}), the complete optimization only needs to consider the constraints in (\ref{eqa:obj_d}-\ref{eqa:obj_f}). Even this, the constrained optimization is typically time-consuming. To address this issue, the MINCO framework \cite{wang2022geometrically} transforms it into an unconstrained optimization problem detailed as follows. First, the time constraints in (\ref{eqa:obj_d}) are equivalent to $T_i > 0$, which can be parameterized as $T_i = e^{\tau_i}, 1 \leq i \leq M$ that always satisfies (\ref{eqa:obj_d}) for $\tau_i \in \mathbb{R}$. Then, the feasibility constraints (\ref{eqa:obj_e}) and (\ref{eqa:obj_f}) can be softly penalized in the cost function by a $C^2$-continuous barrier function \cite{wang2021robust}:
\begin{equation}
 \mathcal L_\mu (x) = \begin{cases}0 & \text { if } x \leq 0, \\ (\mu-x / 2)(x / \mu)^{3} & \text { if } 0<x<\mu, \\ x-\mu / 2 & \text { if } x \geq \mu .\end{cases}
\end{equation}
where $\mu$ is a constant smoothness factor (0.02 in this paper), thus a finite weight for penalty can enforce the constraint at any specified precision. As a consequence, the optimization in (\ref{eq:optimization}) can be turned to an unconstrained form as:
\begin{equation}
 \label{eqa:obj}
 \begin{aligned}
 \min\limits_{\boldsymbol{\tau},\mathbf q} \mathcal J &=\sum\limits_{i=1}^M \left( \int_{0}^{T_i} \| p_i^{(s)}(t) \|_2^2 dt + \rho_T e^{\tau_i} \right) \\
 + &{{\rho_{\mathrm{vel}} \sum\limits_{i=1}^M}\int_{{0}}^{T_i}\mathcal L_\mu\left( 
 \|p^{(1)}_i(t) \|_2^2 - v_{max}^2
 \right) \mathrm dt}\\
 + &{{\rho_{\mathrm{acc}} \sum\limits_{i=1}^M}\int_{0}^{T_i}\mathcal L_\mu\left( 
 \|p^{(2)}_i(t) \|_2^2 - a_{max}^2
 \right) \mathrm dt}\\
 + &{{\rho_{\text{c}} \sum\limits_{i=1}^M}\int_{0}^{T_i}\mathcal L_\mu\left( 
 \|p_i(t) - {o_i} \|_2^2- r_i
 \right) \mathrm dt}
 \end{aligned}
\end{equation}

where $\rho_{\mathrm{vel}},\rho_{\mathrm {acc}},\rho_{\text{c}}$ are the corresponding weight of maximum speed, maximum acceleration and collision-free penalty, and $o_i$ is the center and $r_i$ is the radius of the $i$-th sphere. As shown in \cite{wang2022geometrically}, all gradient of the objective (\ref{eqa:obj}) with respect to waypoints $\mathbf q$ and time allocation $\boldsymbol{\tau}$ can be computed analytically, so a Quasi-Newton method (\ie~LBFGS\footnote{https://github.com/ZJU-FAST-Lab/LBFGS-Lite}) is used to solve the optimization problem effectively.

\section{Planner}

In this section, we present the frontend design that enables high-speed trajectory optimization, which is the main contribution of this paper. 

\subsection{Sphere-Shaped Corridor}
\label{sec:sfc}

\begin{figure}[htpb]
 \centering 
 \includegraphics[width=0.35\textwidth]{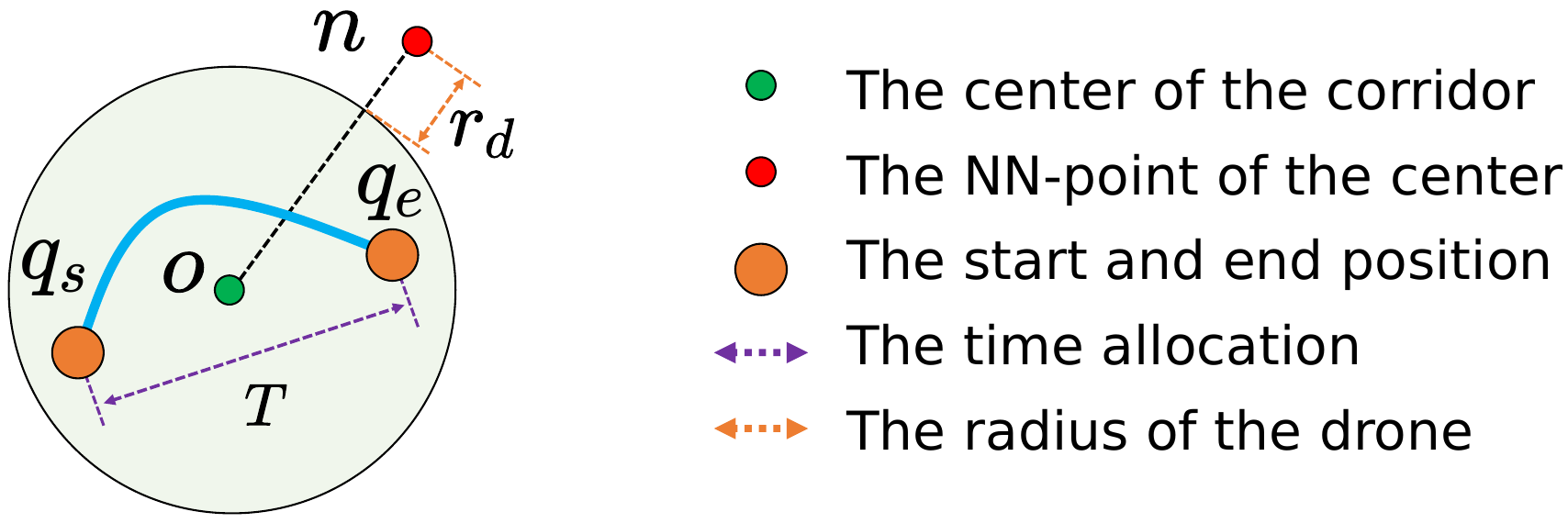}
 \caption{The definition of one sphere and a piece of trajectory in it. $q_s,q_e$ are the start and end point of the trajectory and $T$ is the time allocation. $o$ is the center of the sphere and $n$ is the nearest obstacle point.}
 \label{fig:bubble_def}
\end{figure}

As shown in Fig.~\ref{fig:bubble_def}, a sphere is defined by its center $o\in\mathbb R^3$, the nearest obstacle point $n \in\mathbb R^3$, and the radius:
\begin{equation}
 \label{eqa:radi}
 r = \left\| o - n\right\|_2 - r_d
\end{equation}
where $r_d$ is the radius of the drone. During the trajectory optimization process, each piece of trajectory is constrained in the corresponding sphere to satisfy safety constraints.

To generate a new sphere, we first build a KD-Tree with the obstacle point cloud. Then, for a given center of the sphere $o$, a nearest neighbor search (NN-Search) is performed on that KD-Tree to find the nearest obstacle point $n$, which then determines the radius as in (\ref{eqa:radi}). We call this process \textbf{GenerateOneSphere}$(o)$, which will be used in the sequel.

\subsection{Flight Corridor Generation}
\label{sec:genbub}
The main workflow of the flight corridor generation is described in Alg.~\ref{alg:genbubs}, where a complete flight corridor $\mathcal{B}$ is generated from the given initial position ${p}_0$, goal position ${p}_g$, and a global guide path $\mathcal T$ generated by A*\cite{astar}. The algorithm initializes with a largest possible sphere $\mathcal B_{cur}$ around the initial position ${p}_0$ (Line 2-3). Then, in Line \ref{alg:genbubs:getforward}, a local guide point $p_h$ is selected from the guide path $\mathcal T$, which is the nearest point out of the current sphere $\mathcal B_{cur}$, and a new sphere is generated by \textbf{BatchSample}($p_h, \mathcal{B}_{cur}$) (Sec.~\ref{sec:batch_sample}) and added to $\mathcal B$. This process repeats until the goal position $p_g$ is included in the new generated sphere (Line 8-10).

With the found flight corridor $\mathcal{B}$, the initial waypoint position $\mathbf q$ and time allocation $\mathbf{T}$ are initialized by the function \textbf{WaypointAndTimeInitialization} ($\mathcal B$)(Sec.~\ref{sec:tpallo}) and then optimized in the backend (Sec. \ref{sec:minco}).

\begin{algorithm}[ht]
 \small
 \caption{GenerateCorridorAlongPath()}
 \label{alg:genbubs}
 \textbf{Notation}: The flight corridor $\mathcal B$; global guide path $\mathcal{T}$; Initial and goal position: $ p_0, p_g$; local guide point $p_h$\\
 \KwIn{
 $\mathcal{T}, p_0, p_g$
 }
 \KwOut{
 $\mathcal B$
 } 
 \BlankLine
 \label{alg:genbubs_init}
 Initialize $\mathcal{B}_{cur}$ = GenerateOneSphere($ p_0$)\;
 $\mathcal B$.PushBack($\mathcal{B}_{cur}$)\; 
 \While{\textnormal{True}}{\label{alg:genbubs_while}
 $p_h$ = GetForwardPointOnPath($\mathcal{T},\mathcal{B}_{cur}$)\label{alg:genbubs:getforward}\;

 $\mathcal{B}_{cur}$ = BatchSample($p_h, \mathcal B_{cur}$)\;

 $\mathcal B$.PushBack($\mathcal{B}_{cur}$)\;

 \If(){$p_g \in \mathcal{B}_{cur}$}{break\;}
 }\label{alg:genbubs_end_while}
 
 WaypointAndTimeInitialization($\mathcal B$)\;

\end{algorithm}

\subsubsection{Batch sample}
\label{sec:batch_sample}
 The problem of trajectory optimization under flight corridor constraints is highly non-convex, which means overly conservative constraints may lead to local-minimum or even infeasible solution when the quadrotor initial speed is high. Existing methods \cite{gao2019flying,ji2021mapless} only considered the connectivity between two adjacent spheres. To preserve larger space for the quadrotor to maneuver hence improve the feasibility of the trajectory optimization (\ref{eqa:obj}) at high-speeds, we propose a novel batch sample method to generate a high-quality corridor. We consider this problem in the following aspects: \textbf{(a)} the volume of each sphere: a sphere with larger size can better approximate the real free space with fewer number of spheres, making the optimization problem less constrained, \textbf{(b)} the volume of the overlapped spaces between two adjacent spheres: as discussed in Sec.~\ref{sec:minco}, all waypoints of the trajectory are constrained in the intersecting space, a larger intersecting space means more freedom for the optimization process.

\begin{figure}[ht]
 \centering 
 \includegraphics[width=0.45\textwidth]{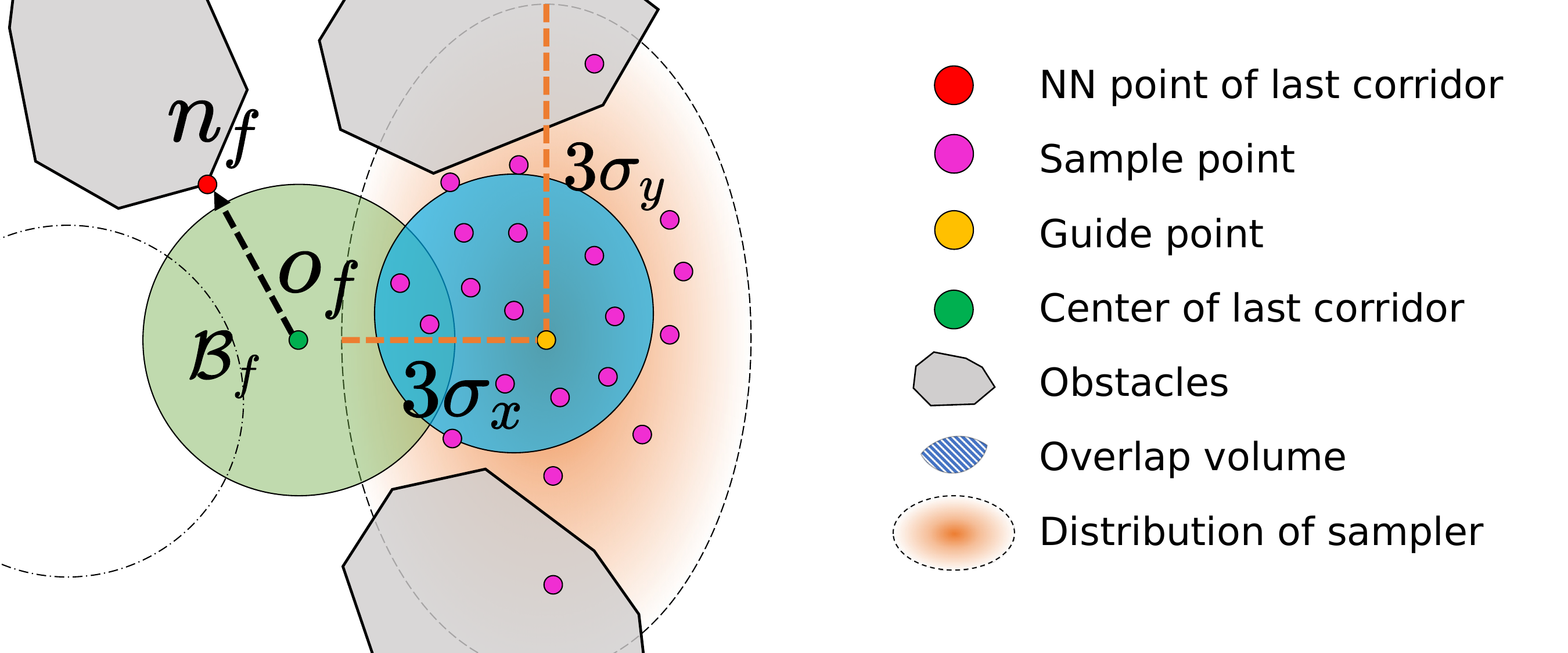}
 \caption{The green circle $\mathcal{B}_f$ is the sphere generated in last round of batch sample. The yellow point is the guide point $p_h$. The purple points are the sampled points according to the probability distribution represented by the orange-shaded area. The blue circle is the best sphere in this round.}
 \label{fig:samp}
\end{figure}

\begin{algorithm}[htpb]
 \caption{BatchSample()}
\small
 \label{alg:batchsample}
 \textbf{Notation}: Last sphere $\mathcal B_f$; Guide point $p_h$; Best sphere in this round $\mathcal B_{best}$; Random sampler $\mathcal S$; Maximum sample num $K$; Safe distance $r_d$; Priority queue sorted by sphere's score: $\mathcal{Q}$;\\
 \KwIn{
 $\mathcal B_f$, $p_h$, $K$
 }
 \KwOut{
 $\mathcal B_{best}$
 } 
 \BlankLine
 Initialize: $\mathcal S$.init($\mathcal B_f$, $p_h$), $k=0$ \label{alg:batchsample_init}\;
 \While{$k<$ $K$}{\label{alg:samp_while}
 $p_{cand}$ = $\mathcal S$.GetOneRandomSample()\;

 $\mathcal B_{cand}$ = GenerateOneSphere($p_{cand}$)\;
 $\mathcal B_{cand}$.score = ComputeScore($\mathcal B_{cand}$)\;
 $\mathcal{Q}$.PushBack( $\mathcal B_{cand}$ )\;
 $k=k+1$\;
 }\label{alg:samp_while_end}
 \If(){$\mathcal{Q}.empty()$}{
 return BatchSampleFailed\;
 }
 $\mathcal B_{best}$ = $\mathcal{Q}$.top()\;
\end{algorithm}

The sampling process is shown in Alg.~\ref{alg:batchsample}. We first initialize the sampler $\mathcal S$ in Line \ref{alg:batchsample_init}. As shown in the orange area of Fig.~\ref{fig:samp}, the sampler generates a random candidate point $p_{cand}\in \mathbb R^3$ under a 3D Gaussian distribution $N(\mu,\Sigma)$, where the mean is set at the guide point $\mu = p_h$ and the covariance is set as $\Sigma = \text{diag} \left(\sigma_x, \sigma_y, \sigma_z \right)$, $\sigma_x = \frac{1}{3}\left\| o_f - p_h \right\|_2, \sigma_z = \sigma_y = 2\sigma_x$, where $o_f$ is the center of last sphere and the $\sigma_x$ direction is aligned with the direction of $o_f-p_h$.

Then in Line \ref{alg:samp_while}-\ref{alg:samp_while_end}, a total number of $K$ points (called a batch) are randomly sampled with $\mathcal S$, each has its score computed by the function \textbf{ComputeScore}($\mathcal B_{cand}$) defined below:
\begin{equation}
 \label{eqa:heu}
 \begin{aligned}
 \text{Score} &= \rho_r V_{\text{cand}} + \rho_v V_{\text{inter}}
 \end{aligned}
\end{equation}
 where $\rho_r,\rho_v \in \mathbb R_+$ are positive weights, $V_{\text{cand}}$ is the volume of the candidate sphere $\mathcal B_{cand}$ and $V_{\text{inter}}$ is the overlapped volume between $\mathcal{B}_{cand}$ and $\mathcal{B}_f$. Finally, the best sphere with the highest score is selected in Line 13.
 
\begin{figure}[t]
 \centering 
 \includegraphics[width=0.4\textwidth]{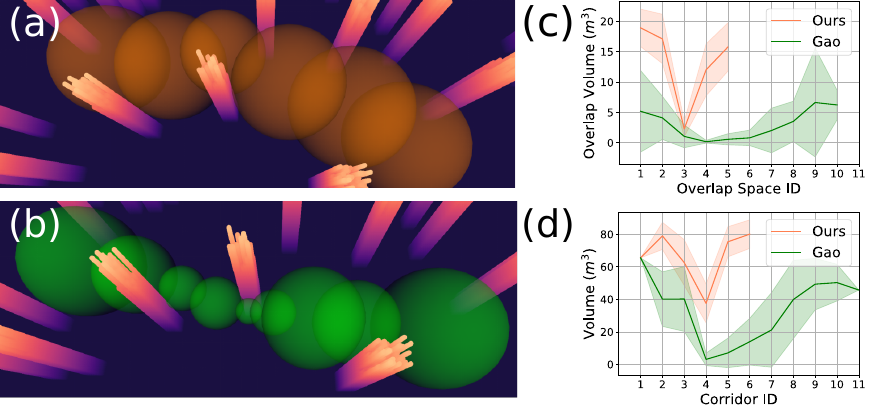}
 \caption{Corridor generation comparison. (a) The corridor generated by our proposed method in one test (b) The corridor generated by Gao \etal~\cite{gao2019flying} in the same test. (c) The comparison of overlapped volume between two adjacent spheres over 100 tests. (d) The volume of each sphere over 100 tests. The shaded area denotes the maximum and minimum value over 100 tests.}
 \label{fig:bub_cmp}
\end{figure}

As shown in Fig.~\ref{fig:bub_cmp}, compared with Gao \cite{gao2019flying}, the proposed method can better approximate the real free space with fewer spheres and larger sphere sizes. Furthermore, our algorithm has lower computational complexity than Gao's approach, which uses an RRT-like method and takes samples from the whole space. Our process follows a coarse-to-fine manner, where we first use A* to find the shortest path and then take batch samples only around this path. In this way, the sample space, hence computation time, is significantly reduced. We test 100 times in the same environment shown in Fig.~\ref{fig:bub_cmp}. The proposed method only takes an average $0.74~ms$ to generate the corridor, while Gao's method takes an average $100~ms$.

\subsubsection{Waypiont and Time Initialization}
\label{sec:tpallo}

\begin{figure}[htbp]
 \centering 
 \includegraphics[width=0.4\textwidth]{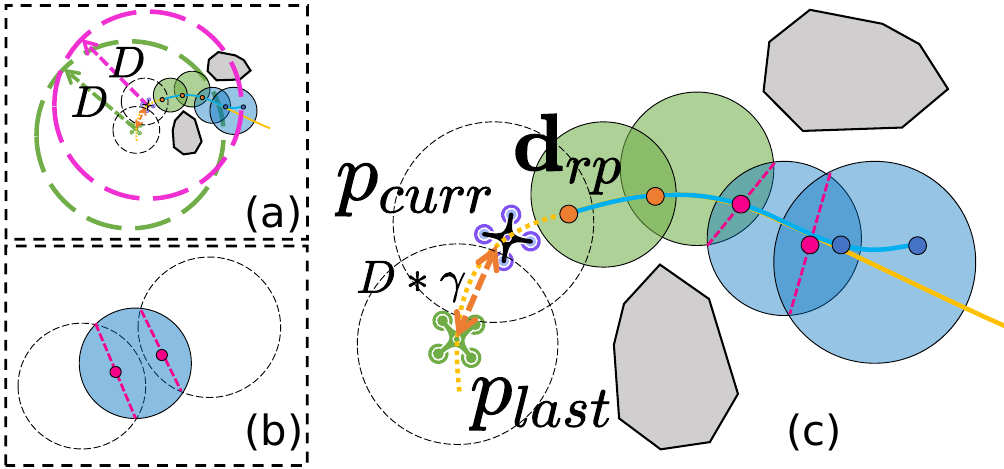}
 \caption{The receding horizon corridors strategy. (a) The green and pink dashed circle are respectively the planing horizon in last and current replan. (b) The pink point is the center of the overlap area, which is used by the \textit{Default Initialization}. (c) Spherical corridor in green are previously generated, using the \textit{Hot Initialization}. And corridor in blue are newly generated, using the \textit{Default Initialization}.}
 \label{fig:rhp}
\end{figure}

For a given flight corridor $\mathcal{B}$, we adopt a \textit{Default Initialization} strategy, where the waypoint are initialized as the center of the overlap space between two adjacent spheres (pink points in Fig.~\ref{fig:rhp}(b)), and the time allocation is initialized as $T_i = \frac{\left\| q_{i} - q_{i-1}\right\|_2}{v_{max}}$.

\subsection{Receding Horizon Corridors in Replan}
\label{sec:rhc}
During a high-speed flight in an unknown environment, the quadrotor needs to replan frequently to avoid newly sensed obstacles. We use a distance-triggering replaning strategy. Specifically, the trajectory is planned (both frontend corridor generation and backend optimization) in a fixed distance $D$ (\ie~planning horizon) depending on the sensing range. Denote the position of last replan as $p_{last}$ and current quadrotor position as $p_{curr}$. The replan process is triggered if $\left\|p_{last} - p_{curr}\right\|_2>\gamma \cdot D$, where $\gamma \in [0,1]$ is a constant ratio. In this way, as the drone moves forward, the newly sensed obstacle can be actively handled by the replan process. A replan is also triggered when the current trajectory under execution is found to collide with any obstacles. 

A major challenge in the replan occurs when the quadrotor speed is high, which requires sufficient space for the quadrotor to maneuver such that the newly sensed obstacles can be avoided successfully. Corridor generation without considering the quadrotor's current state \cite{liu_sfc,gao2019flying,ji2021mapless} often causes too small feasible region in the trajectory optimization (\ref{eq:optimization}), which is difficult (or even impossible) to solve (\eg, by optimizing (\ref{eqa:obj})). Another problem is that with the increase of the current speed, the objective function becomes highly non-convex. As described in Sec.~\ref{sec:minco}, our optimization problem is turned into an unconstrained one. The non-convexity of the objective function may cause the optimization with the \textit{Default Initialization} to easily stuck at a bad local minimum which violates the collision-free or kinodynamic constraints.

We solve these problems by a \textit{Receding Horizon Corridors} (RHC) strategy shown in Fig.~\ref{fig:rhp}. The key is to reuse a few spheres from the previous planning cycle in current replan. Concretely, when a new replan is triggered, the nearest future waypoint $\mathbf d_{rp}$ in $\mathbf{q}$ is selected as the initial state. A few spheres after $\mathbf d_{rp}$ will be reused to constitute the first part of the new corridor, followed by newly generated spheres reaching the current planning horizon $D$. This receding scheme ensures the corridor in each replan always contains sufficient space for the quadrotor to maneuver from its current state (since the current quadrotor state is on the previous trajectory, which is contained in the previous corridor), hence significantly enlarging the feasible region in the backend trajectory optimization. In experiments, we reuse spheres that fall within a certain distance (\eg, $3m$) of the current quadrotor position $p_{curr}$. Furthermore, to speed up the trajectory optimization and mitigate the local minimum issue, the waypoints $\mathbf q$ and time allocation $\mathbf T$ contained in the reused corridor, which were optimized in the previous planning cycle, are used to initialize the current trajectory optimization (\ie~\textit{Hot Initialization}). The waypoints and time allocation in the newly generated spheres are still initialized by the default scheme (Sec. \ref{sec:tpallo}).

\section{Experiments} 
\subsection{Benchmark Comparison}
\label{sec:bench}

\begin{figure}[htbp]
 \centering 
 \includegraphics[width=0.4\textwidth]{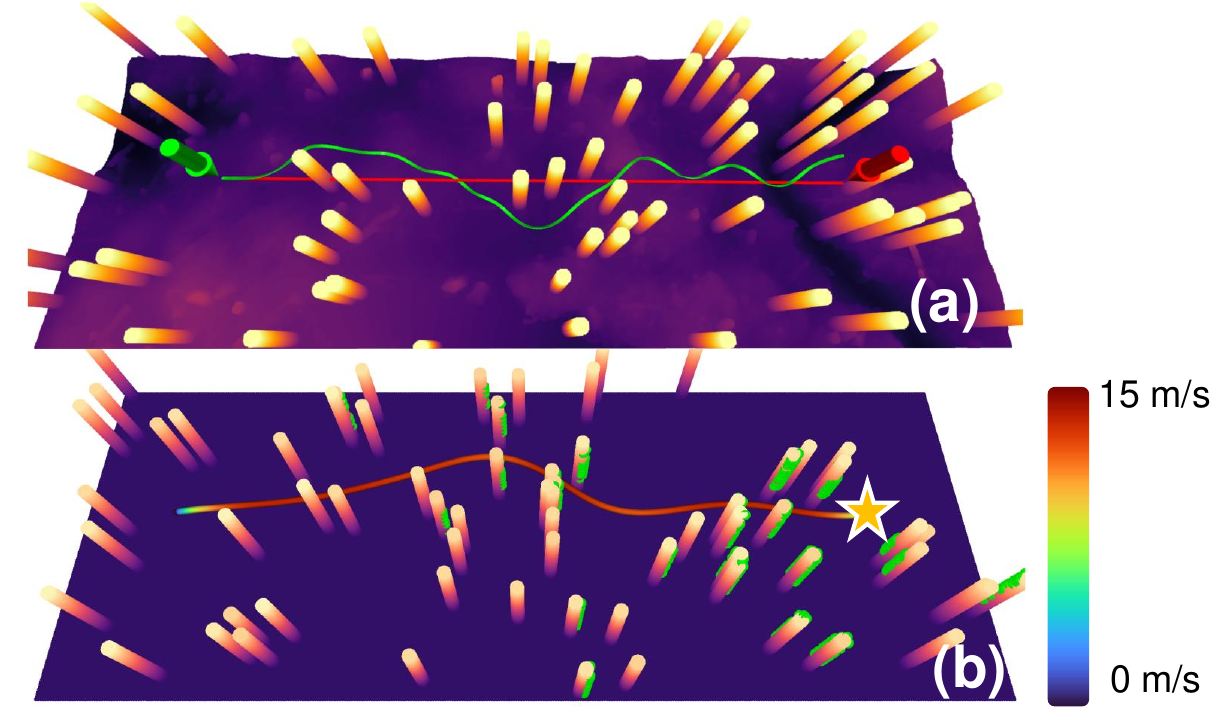}
 \caption{(a) The executed trajectory in Loquercio \etal~\cite{loquercio2021learning}. (b) The executed trajectory with the proposed method is colored with forward speed from $0~m/s$ to $15~m/s$. The yellow star is the initial position of the drone, and the green points are the simulated LiDAR points.}
 \label{fig:limit}
\end{figure}
In this section, we compare the proposed method with a most recent planning work based on imitation learning \cite{loquercio2021learning} (Learning), and two model-based planning methods evaluated by it, including a frontend-backend type optimization-based method from Zhou \etal~\cite{zhou2019robust} (FastPlanner) and a reactive planner designed for the high-speed flight from Florence \etal~\cite{florence2020integrated} (Reactive). We evaluate the performance of our method in a simulated forest environment used by the learning method. Due to the unavailability of the simulation environment used by the original work \cite{loquercio2021learning}, we reproduce the environment according to their description. Specifically, the forest has trees distributed in a rectangular region $R(l,w)$ of width $w$ and length $l$, the origin lies in the center of $R$. Trees are randomly placed according to a homogeneous Poisson point process $P$ with the intensity $\delta~tree/(m^2)$. The sensor input in the simulation includes a simulated LiDAR point cloud, with the sensing range of $8~m$ at $30~Hz$ (see green points in Fig.~\ref{fig:limit}(b)). The quadrotor full state is assumed to be known to eliminate the influence of state estimation. 

\begin{figure}[htbp]
 \centering 
 \includegraphics[width=0.4\textwidth]{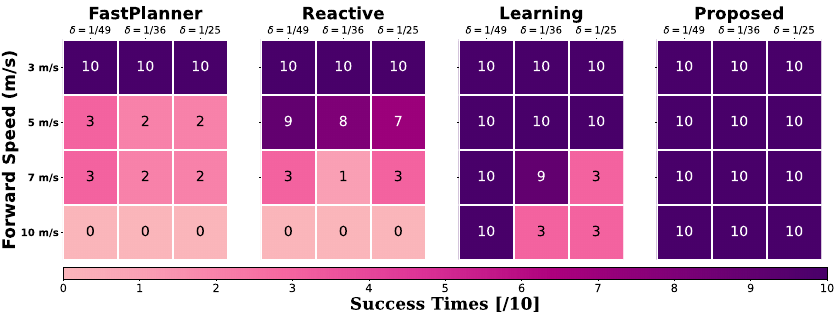} 
 \caption{The success rate comparison for different methods. The proposed method keeps a high success rate in all simulated test environments with varying forest densities.}
 \label{fig:bench}
\end{figure}

\begin{figure*}[t]
 \centering 
 \includegraphics[width=1.0\textwidth]{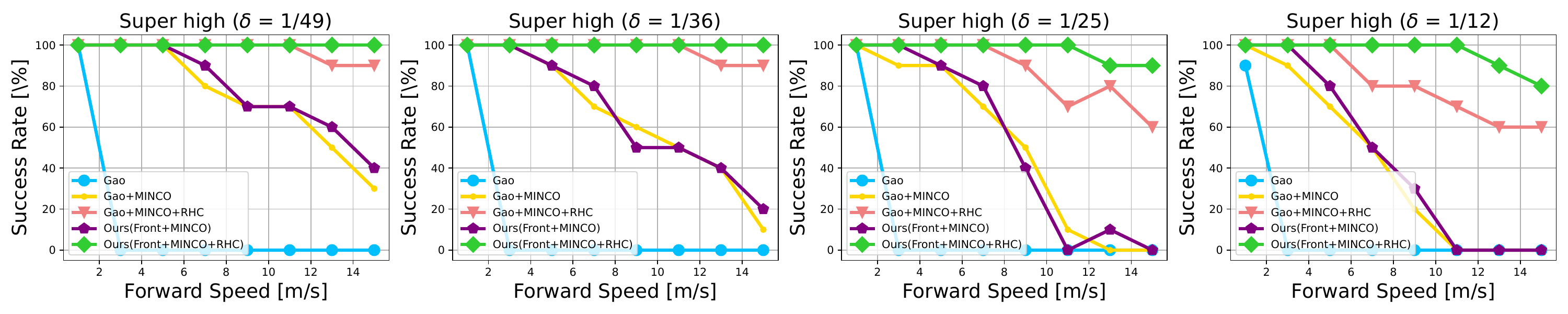}

 \caption{The ablation study of the proposed method. The green line is the proposed method (\textit{Ours(Front+MINCO+RHC)}). The blue line (\textit{Gao}) is the original version of Gao \etal's work \cite{gao2019flying}. The yellow line (\textit{Gao+MINCO}) uses corridor generation from \cite{gao2019flying}, but with trajectory optimization replaced by MINCO in Sec.\ref{sec:minco}. The purple line (\textit{Ours(Front + MINCO)}) is our method without the RHC strategy. The red line (\textit{Gao+MINCO+RHC}) uses Gao \etal's corridor generation method and the same MINCO optimization and RHC strategy as ours.
 }
 \label{fig:ablation}
\end{figure*}

We use exactly the same configuration in \cite{loquercio2021learning} to make a fair comparison: $w = 30~m$ and $l = 60~m$, and the start zone of the drone is at $(-{l}/{2}, 0)$, the goal position $({l}/{2}, 0)$. Three different tree densities with $\delta = {1}/{49}$ (low), $\delta = {1}/{36}$ (medium), and $\delta = 1/25$ (high) are tested. {In each experiment, we use different random seed to generate different simulated maps.} One flight is considered to be successful only if the drone reaches the goal without violating the velocity, acceleration, or collision-free constraints. The results are shown in Fig.~\ref{fig:bench}. Similar to \cite{loquercio2021learning}, we test our method 10 times in each different density or speed and compute the success rate of each, and the results of other baseline are directly obtained from \cite{loquercio2021learning}. Noting that in \cite{loquercio2021learning}, the maximum mass-normalized thrust of the simulated drone is limited to $35.3~m/s^2$, while we limit our simulated drone to $15~m/s^2$. As can be seen, our approach outperforms others in all cases, even with a lower thrust limit. Moreover, compared with Loquercio \etal~\cite{loquercio2021learning}, the proposed method generates much smoother trajectories, which is usually easier to track (see Fig.~\ref{fig:limit}).

\subsection{Ablation Study}
To further validate each module of the proposed method, we compare our method in detail with Gao \etal~\cite{gao2019flying}, which generates sphere-shaped corridors in an RRT* style and optimizes a minimal snap trajectory with fixed time allocation. We use the same simulated map configuration mentioned in Sec.\ref{sec:bench}, but further add tests with $\delta = 1/12$ (super high). The key three elements of our approach includes the trajectory optimization in Sec.~\ref{sec:minco} (MINCO), the frontend corridor generation in Sec.~\ref{sec:genbub} (Front), and the receding horizon corridors strategy (RHC) in Sec.~\ref{sec:rhc}. A series of ablation studies are performed, and the results are shown in Fig.~\ref{fig:ablation}. \textit{Gao} is the original version from \cite{gao2019flying}. This method fails to generate trajectory with speed over $2~m/s$ due to the inability to optimize time allocation in the backend. To fix this issue, we replace the backend of \textit{Gao} by MINCO (\textit{Gao+MINCO}) and compare it with our method without RHC strategy (\textit{Ours (Front+MINCO)}). The performances of the two {are} very close, showing that {MINCO} can generate more aggressive trajectories and that our frontend alone does not improve the success rate much. Furthermore, we incorporate the RHC strategy to the method \textit{Gao} (\textit{Gao + MINCO + RHC}) and compare it with our full algorithm (with both frontend and RHC). As can be seen, each method with RHC has a significantly higher success rate at high speeds on all map densities, verifying the effectiveness of the RHC strategy. Moreover, our full algorithm with our frontend (\textit{Ours(Front + MINCO + RHC)}) achieves a higher success rate than \textit{Gao} with the same MINCO and RHC strategy (\textit{Gao+MINCO+RHC}), showing the effectiveness of our frontend in the overall planning system.

\subsection{Run Time Analysis}
In this section, we compare the run time of the proposed method with the baseline. We test our method both on the desktop computer, with a 2.90 GHz Intel i7-10700 CPU, and an onboard computer with a 1.1 GHz Intel i7-10710U CPU. The baseline FastPlanner \cite{zhou2019robust} and Gao \cite{gao2019flying} are tested on the same desktop computer. The test environment is a simulated forest with $\delta = \frac{1}{25}$ shown in Fig.~\ref{fig:limit}(b). The computation time is divided into two parts: mapping and planning. For FastPlanner, the mapping process includes building a Euclidean signed distance field (ESDF), and planning includes frontend path-search and backend trajectory optimization. For Gao's method, the mapping process includes a static KD-Tree update, and the planning includes corridor generation and SOCP optimization. For the proposed method, the mapping includes the update of an OctoMap \cite{hornung2013octomap} (no ray-casting) and an incremental KD-Tree (\ie, ikd-tree \cite{cai2021ikd}). The planning includes frontend A* search, corridor generation, and trajectory optimization. As shown in Table~\ref{tab:run_time}, the proposed method enjoys much lower computational complexity, which can replan at over $50~Hz$ even on the onboard platform.
\begin{table}[htbp]
 \small
 \centering
 \caption{Run Time Comparison}
 \setlength{\tabcolsep}{1mm}
 \begin{tabular}{@{}ccccccc@{}}\toprule
 \label{tab:run_time}
 Method &Components & $\mu$ [ms] & $\sigma$ [ms] & Total [ms]\\
 \hline
 \multirow{2}{*}{Fast-Planner \cite{zhou2019robust}} 
 &Mapping & 38.20 & 6.90 &\multirow{2}{*}{42.92} \\
 &Planning& 4.72 & 1.60\\ \hline 
 \multirow{2}{*}{Gao \cite{gao2019flying}} 
 &Mapping & 12.78 & 3.69 &\multirow{2}{*}{167.01} \\
 &Planning& 154.23 & 40.60\\ \hline 
 \multirow{2}{*}{\textbf{Ours}} 
 &Mapping& 3.16 & 0.76& \multirow{2}{*}{\textbf{4.69}} \\
 &Planning & 1.53 & 0.63 \\ \hline
 
 \multirow{2}{*}{Ours (onboard)} 
 &Mapping &8.97 & 6.51& \multirow{2}{*}{13.31}\\
 &Planning&4.34 &2.33\\ \hline
 
 \end{tabular}
 \end{table}

 \subsection{Real-world Experiments}

\begin{table}
 \scriptsize
 \centering
 \caption{Detailed Profile of 11 Real-world Tests}
 \setlength{\tabcolsep}{1mm}
 \begin{tabular}{@{}ccccccc@{}}\toprule
 \label{tab:real_trajs}
 & Executing time [s] & Length [m] & Average Vel. [m/s]& Max Vel.[m/s]\\
 \hline
Test1&15.59&111.18&7.05&8.02\\
Test2&5.81&41.61&6.97&\textbf{13.72}\\
Test3&5.69&40.65&6.92&12.00\\
Test4&6.16&34.63&5.47&8.80\\
Test5&17.21&79.17&4.54&5.01\\
Test6&9.90&42.75&4.23&6.54\\
Test7&9.60&58.65&6.01&7.00\\
Test8&6.99&37.55&5.21&7.05\\
Test9&6.02&45.72&7.34&11.64\\
Test10&5.70&29.39&5.01&7.00\\
Test11&5.40&45.08&\textbf{8.11}&12.00\\ \hline
 \end{tabular}
 \end{table}

\begin{figure}[t]
 \centering 
 \includegraphics[width=0.4\textwidth]{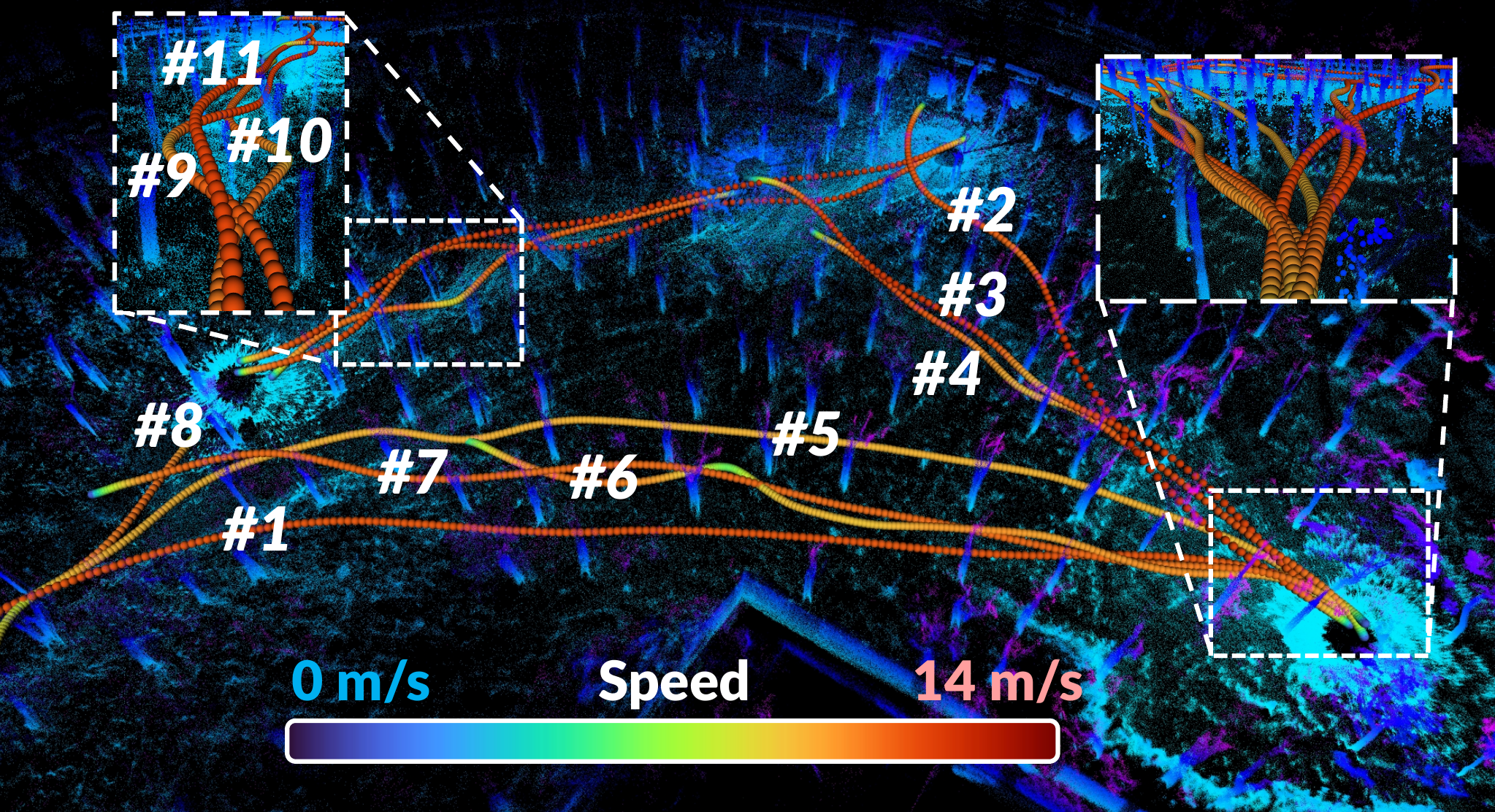}
 \caption{Composite image of 11 real-world flight trajectories colored by their speed. Each experiment (trajectory) is conducted independently with real-time mapping. After all experiments, the executed trajectory along with the map built during each flight are registered together to produce this composite image. The tree crown are removed to better show the trajectories.}
 \label{fig:all_real_traj}
 \end{figure}

 To verify our planning method in real-world environments, we build a LiDAR-based quadrotor platform. The platform has a total weight of $1.45~kg$ and can produce a maximum thrust over $60~N$, resulting in a thrust-to-weight ratio of $4.1$. For localization and mapping, we use the Livox Mid360 LiDAR and PixHawk flight controller's built-in IMU running FAST-LIO2 \cite{xu2022fast} (the sensors are initialized by LI-Init\cite{zhu2022robust}), which provides 100 $Hz$ high-accuracy state estimation and $25~Hz$ point cloud. The trajectory tracking controller is an on-manifold model predictive controller in \cite{lu2021model}, the planning horizon is set to $D = 15~m$ and replan ratio $\gamma = 0.4$. All perception, planning, and control algorithm are running on an Intel NUC with CPU i7-10710U in real-time. We have done 12 experiments in a forest environment with maximal speed ranging from $5~m/s$ to $14~m/s$. All the experiments succeeded except one due to a controller failure. Fig. \ref{fig:all_real_traj} shows the experimental environment and all the trajectories colored by their speed. As can be seen, our planner is robust by accomplishing all the tests in the real-world environment. Fig. \ref{fig:fig1} shows the third person of the quadrotor in one flight. More quantitatively, Table~\ref{tab:real_trajs} summarizes the detailed trajectory profiles including the trajectory executing time, total length, average and maximum speed. As can be seen, our method achieves an average speed up to $8.11m/s$ and a maximum speed of $13.7m/s$. To the best of our knowledge, this is the highest speed that a fully autonomous quadrotor can achieve in a real-world, cluttered, and unknown environment (see Fig. \ref{fig:real_vel}). More visual illustration of the experiments is shown in our video\footnote{\href{https://youtu.be/7tQCV6KBzSY}{https://youtu.be/7tQCV6KBzSY}}. 
 
\begin{figure}[t]
 \centering 
 \includegraphics[width=0.45\textwidth]{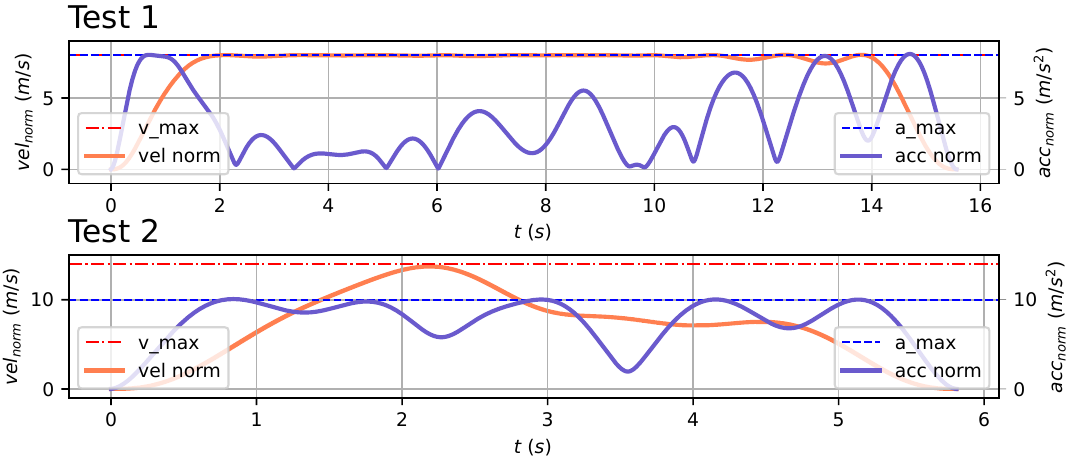}
 \caption{The norm of velocity and acceleration. The executed trajectory of \textit{Test 1} is about $111.18~m$, and \textit{Test 2} is about $41.61~m$. The average speed is about $7~m/s$ for both test, and the maximum speed is over $13.7~m/s$ in \textit{Test 2}.}
 \label{fig:log}
 \end{figure}

Fig.~\ref{fig:log} shows the speed and acceleration profiles of two typical flights, called \textit{Test 1} (long trajectory length) and \textit{Test 2} (high flight speed). For \textit{Test 1}, we limit the maximum speed to $8~m/s$ and the maximum acceleration to $8~m/s^2$. In \textit{Test 2}, a more agile flight is performed where the maximum speed is $14~m/s$ and the maximum acceleration is $10~m/s^2$. As can be seen, both speed and acceleration constraints are well satisfied. 

\section{Conclusion and Future Work}

In this paper, we propose a novel motion planning algorithm that generates smooth, collision-free, and high-speed trajectories in real-time. The whole planning system can work with fully onboard sensing, and computation at a replan frequency over $50~Hz$. To enable high-speed flight in the wild, we proposed two novel designs. One is a sampling-based sphere-shaped corridor generation method, which can generate high-quality corridors (\ie~larger size and bigger overlaps) in a relatively short time. Another is a \textit{Receding Horizon Corridors} strategy, which fully utilizes previously generated corridors and the optimized trajectory. With these designs, the proposed method significantly increases the replan success rate in high-speed cases.

One limitation of our algorithm is that the reused corridors from last planning cycle are not guaranteed to be obstacle-free due to newly sensed obstacles that may be occluded in previous LiDAR measurements. This will cause the reused corridor to be discarded and hence occasionally lower the success rate when the environment is extremely cluttered. This limitation can be overcome by placing the first few corridors of a (re-)plan in known free spaces (instead of free and unknown spaces), so that these free corridors can be safely reused in the next planning cycle. Restraining the first few spheres in free spaces also enables the planning of a safe backup trajectory like \cite{tordesillas2021faster} which guarantees a safe flight. In the future, we will explore these designs and extend the method to more different missions and environments.

\section*{Acknowledgment}
The authors gratefully acknowledge DJI for fund support and Livox Technology for equipment support during the whole project. The authors would like to thank Guozheng Lu and Wei Xu for the helpful discussions.

{\small
\bibliographystyle{unsrt}
\bibliography{main}
}

\end{document}